\documentclass{sig-alternate}
\usepackage{amsmath}
\usepackage{amsfonts}
\usepackage{amssymb}
\usepackage{times}
\usepackage{microtype}
\usepackage{color}
\usepackage{url}
\usepackage{algorithm2e}
\usepackage{caption}

 \usepackage{upgreek}

\newcommand{\order}[1]{\textit{O}(#1)}

\title{Scalable Modeling of Conversational-role based Self-presentation
Characteristics in Large Online Forums}
\numberofauthors{6}
\author{
\alignauthor
Abhimanu Kumar \\
\affaddr{School of Computer Science}\\
\affaddr{Carnegie Mellon University}\\
\email{abhimank@cs.cmu.edu}
\alignauthor
Shriphani Palakodety \\
\affaddr{School of Computer Science}\\
\affaddr{Carnegie Mellon University}\\
\email{shriphanip@gmail.com}
\alignauthor
Chong Wang \\
\affaddr{School of Computer Science}\\
\affaddr{Carnegie Mellon University}\\
\email{chongw@cs.cmu.edu}
\alignauthor
\and
Carolyn P. Rose\\
\affaddr{School of Computer Science}\\
\affaddr{Carnegie Mellon University}\\
\email{cprose@cs.cmu.edu}
\alignauthor
Eric P. Xing\\
\affaddr{School of Computer Science}\\
\affaddr{Carnegie Mellon University}\\
\email{epxing@cs.cmu.edu}
\alignauthor
Miaomiao Wen\\
\affaddr{School of Computer Science}\\
\affaddr{Carnegie Mellon University}\\
\email{mwen@cs.cmu.edu}
}

\begin{document}
\maketitle
\begin{abstract}

Online discussion forums are complex webs of overlapping subcommunities 
(macrolevel structure, across threads) in which users enact different roles 
depending on which subcommunity they are participating in within a particular 
time point (microlevel structure, within threads).  This sub-network structure 
is implicit in massive collections of threads. To uncover this  structure, 
we develop a scalable algorithm based on stochastic variational inference and 
leverage topic models (LDA) along with mixed membership stochastic block (MMSB) 
models. We evaluate our model on three large-scale datasets, Cancer-ThreadStarter (22K
users and 14.4K threads), Cancer-NameMention(15.1K users and 12.4K threads) and
StackOverFlow (1.19 million users and 4.55 million threads). Qualitatively, we
demonstrate that our model can provide useful explanations of microlevel
and macrolevel user presentation characteristics in different communities
using the topics discovered from posts. Quantitatively, we show that our model
does better than MMSB and LDA in predicting user reply structure within threads. 
In addition, we demonstrate via synthetic 
data experiments that the proposed active sub-network discovery
model is stable and recovers the original parameters of the experimental
setup with high probability.

\end{abstract}


\section{Introduction}
Online forums are a microcosm of communities where users' presentation
characteristics vary across different regions of the forum. Users participate in
a discussion or group activity by posting on a related thread. During his
stay on a forum, a user participates in many different discussions and posts on
multiple threads. The thread level presentation characteristics of a user are different
than the global presentation characteristics. A participating user gears his
responses to suit specific discussions on different threads. These thread based
interactions give rise to active sub-networks, within the global network of users,
that characterize the dynamics of interaction. Overlaying differential changes
in user interaction characteristics across these sub-networks provides
insights into users' macroscopic (forum-wide) as well as microscopic (thread
specific) participation behavior. 

Analysing online social networks and user forums have been approached using
various perspectives such as network ~\cite{Shi:2000:NCI:351581.351611,
Shi00learningsegmentation} , probabilistic 
graphical models~\cite{ Airoldi:2008:MMS:1390681.1442798}, 
combined network \& text 
~\cite{Ho:2012:DHT:2187836.2187936,Nallapati:2008:JLT:1401890.1401957}. 
However none of these have taken
into account the dynamics of sub-networks and the related thread-based framework
within which forum discussions take place. Whereas
active sub-network modelling has been very useful to the research in
computational biology in recent years where it's been used to model sub-networks
of gene interactions~\cite{journals/ploscb/DeshpandeSVHM10,Lichtenstein:Charleston},
very few approaches using active sub-network have been proposed to model online
 user interactions. Taking into account sub-network interaction
dynamics is important to correctly model the user participant behavior. For
example, users post their
responses on discussion threads after reading through responses of
other users in the threads. The users possibly post multiple times on the thread
 as a form of reply to other posts in the thread. For analysing such
 interactions it becomes imperative that the structure of the conversation must also be taken
into account  besides the user interaction network and the
text posted. This enables us to gain deeper insights into user behavior in the
online community that was not possible earlier. 

One of the main challenges of this work has been the ability to model
active sub-networks in a large forum with millions of users and
threads. A social network spanning around millions of users and threads would 
be an ideal case to demonstrate the effectiveness of sub-network
modelling.
To efficiently scale our model, we derive a scalable inference based on
stochastic variational inference (SVI) with sub-sampling~\cite{Hoffman:2013:SVI} 
that has the capacity to deal with such massive scale data and parameter space.
 The scalability of the SVI
with sub-sampling is further boosted by employing Poisson distribution 
to model edge weights of
the network. A Poisson based scheme need not model zero edges
(\cite{Kerrer:Newman}), where as MMSB style approaches~\cite{Airoldi:2008:MMS:1390681.1442798}
 must explicitly model them.
A further set of
parallelization in inner optimization loops of local variational parameters
pushes the learning speed even more. This work is to date the largest 
modelling of any social graph that also takes user contents into
account. 

\paragraph{Contributions}
\begin{itemize}
  \item This work provides novel insights
into how users' self- representational characteristics vary
depending on the discussion they are in. This is achieved via active sub-network
modelling.
\item Our model outperforms LDA and MMSB in link prediction across three
datasets demonstrating the leverage it gains by combining the two along with discussion
structures in modelling sub-networks.
\item It is highly scalable and is able to achieve convergence in
matter of hours for users and threads that are an order of a million despite the
time complexity of \order{users$\times$users$\times$threads}.
\item  Stability
is another aspect of the proposed new approach and is demonstrated by
the model recovering back its parameters in synthetic experiments. 
\end{itemize}

\section{User Role Modelling}
\label{sec:approach}
Online forums have a specific discussion structure  that
provides a lot of context to all the interactions occurring among the users.
Here we describe a typical forum discussion scenario.

\subsection{Discussion structure in online forums}
When two users interact in a thread or through a post
they play certain conversational roles and project their specific identity.
It is valuable to know what conversational roles each plays (which topic or
community they each belong to) in that interaction. When a user $u$ is
representing community $c$ out of all the communities that he is part of and
participates in a discussion, he tailors his post content accordingly to suit
the explicit or implicit community and discussion norms. Knowing the style of
community specific text content provides a lot of information about that community 
in general. It also provides information about what role user $u$ plays when he
is in community $c$ and engaged in a specific discussion thread $t$. In online
forums multi-user interactions occur a lot i.e.
in a thread a user can post by addressing another specific user but he is
also addressing other users in the thread explicitly or implicitly (via either
gearing his replies to address other users' concerns into consideration or
addressing them directly in the post).
Modeling this phenomenon would bring the model closer to realities of online discussions.
This can be modelled by aggregating users posts across a thread,
though not across the whole of the forum. We will elaborate on this more in the
generative story of our model. 


\subsection{Graphical model \& generative story}
\label{sec:gen-story}
Based on the description above our graphical model is designed as shown
in Figure~\ref{fig:finalThreadAggregationModel}. In this model
we aggregate the posts of 
a given user in a given thread $t$ into one document which has token
set $N_{t,p}$ . This helps us incorporate the knowledge that a user's post is
influenced by the posts of other users present on the thread, assuming that
he reads at least some of them.

The generative process for the model is as follows:

\begin{figure}
\centering
\includegraphics[height=4cm,width=8.5cm]{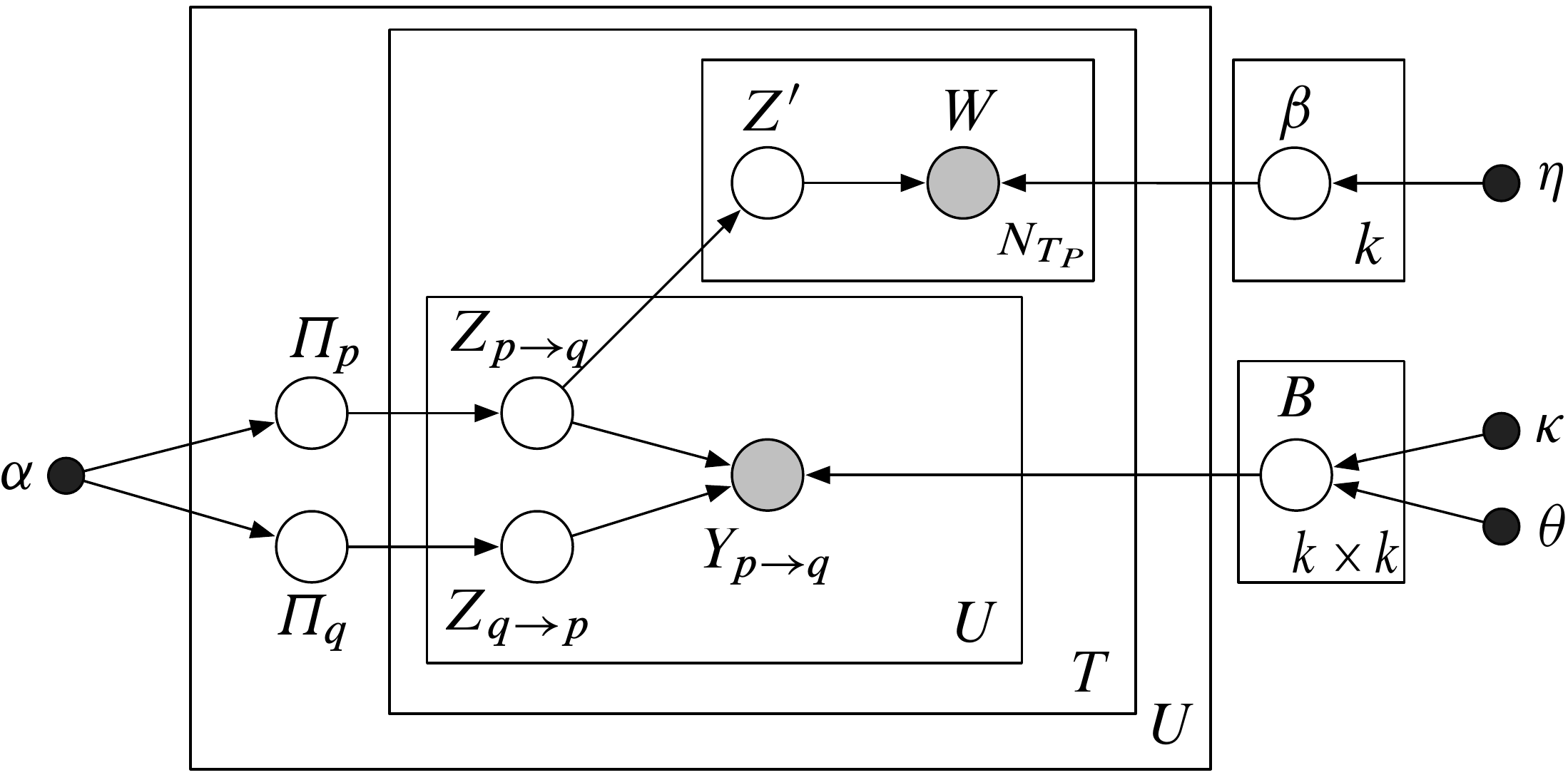}
\caption{The proposed approach models active sub-network of users in the forum.
$U$ si the total number of user in all of the forum, $T$ si the number of
threads. $N_{T_p}$ is the total number of tokens user $p$ posted in thread $t$.
}
\label{fig:finalThreadAggregationModel}
\end{figure}

\begin{itemize}
  \item For each user $p $,
  \begin{itemize}
    \item Draw a $K$ dimensional mixed membership vector 
    $\overset{\rightarrow}{\uppi}_{p} \sim$ Dirichlet($\alpha$).
  \end{itemize}
  \item for each topic pair $g$ and $h$,
    \begin{itemize}
    \item Draw $B(g,h) \sim Gamma(\kappa,\eta)$; where $\kappa, \eta$ are
    parameters of the gamma distribution.
  \end{itemize}

  \item For each pair of users $(p, q)$ and each thread $t$,
  \begin{itemize}
    \item Draw membership indicator for the initiator, 
    $\overset{\rightarrow}{z}_{(p \rightarrow q,t)} \sim$
    Multinomial($\uppi_{p}$).
    \item Draw membership indicator for the receiver,
    $\overset{\rightarrow}{z}_{(p \leftarrow q,t)} \sim$
    Multinomial($\uppi_{q}$).
    \item Sample the value of their interaction, $Y(p,q,t) \sim$
    Poisson( ${\overset{\rightarrow}{z}}^{\top}_{(p \rightarrow q,t)}
    B~\overset{\rightarrow}{z}_{(p \leftarrow q,t)}$ ). 
	\end{itemize}
	\item For each user $p \in t$,
	\begin{itemize}
	  \item Form the set $\delta_{t,p}$ that contains all the users that p
	  interacts to on thread $t$,
	  \begin{itemize}
	    \item For each word $w \in N_{t,p}$, 
	    \item Draw $z^{'}_{t,p,w}$ from $Dirichlet(\sum_{\forall q\in \delta_{t,p}} z_{(t,p
	    \rightarrow q)})$.
	    \item Draw $w \sim \phi(w|\beta,z^{'}_{t,p,w}) $.
	  \end{itemize}
  \end{itemize}
\end{itemize}  

The use of Poisson distribution for $Y(p,q,t) \sim$ \\
    Poisson(${\overset{\rightarrow}{z}}^{\top}_{(p
\rightarrow q,t)} B~\overset{\rightarrow}{z}_{(p \leftarrow q,t)}$) (the 
network edge between the user's $p$ and $q$) besides modelling non-binary edge
strength enables the model to ignore non-edges between users ($Y_{t,p,q}$) and
thus achieve faster convergence~\cite{Kerrer:Newman}. In MMSB style community block models, 
	 non-edges are to be modelled
explicitly.
\begin{equation}
\log L = \log \! P(Y, W, Z_{\leftarrow}, 
Z_{\rightarrow}, \Pi, B, \beta | \alpha, \eta, \theta, \alpha). \\
\end{equation}
The log-likelihood of the model described in \ref{sec:gen-story} is given
 above and derived in detail in the appendix ~\ref{eqn:LL}.

\begin{align}
q = &\prod_{p}q(\Pi_{q} | \gamma_{p}) \prod_{t} \bigg[ \prod_{p, q} \! 
q(Z_{t, p \rightarrow q}, Z_{t, p \leftarrow q} | \phi_{t,p,q})  \nonumber\\ 
\cdot &\prod_{p \in t} \prod_{i=1}^{N_{T_{p}}} q(Z'_{t,p,i} | \chi_{t,p,i})
\bigg] \nonumber \\
\cdot & \prod_{g,h} q(B_{g,h} | \nu_{g,h} \lambda_{g,h}) \prod_{k} q(\beta{k} |
\tau_{k}).
\label{eqn:variationalQ}
\end{align}
We use variational approximation to maximize log-likelihood.
Equation~\ref{eqn:variationalQ} above is the approximation of the log-likelihood and we
use structured mean field~\cite{Xing_et_al:2003} to maximize parameters of $q$.
The local variational parameters, $\phi$ (MMSB parameters) and $\chi$ (LDA)
parameters, are maximized using equations \ref{eqn:phiUp} and \ref{eqn:chiUp}
where $\Delta_{\phi^{'}_{t,p,g,h}}$ and $\Delta_{\chi^{'}_{t,p,g,h}}$ are
defined by equations ~\ref{eqn:phiDelta} and~\ref{eqn:chiDelta} respectively.

\begin{align}
\phi_{t,p,g,h} \propto e^{\Delta_{\phi^{'}_{t,p,g,h}}}.
\label{eqn:phiUp}
\end{align}

\begin{align}
\chi_{t,p,i,k} \propto e^{\Delta_{\chi^{'}_{t,p,g,h}}}.
\label{eqn:chiUp}
\end{align}

The traditional variational updates for global parameters $\gamma, \nu, \lambda$
(MMSB) and $\tau$ (LDA) are defined using equations
\ref{eqn:gammaUp}, \ref{eqn:nuUp}, \ref{eqn:lambdaUp} and \ref{eqn:tauUp}
respectively (details are in the appendix).

\paragraph{Terminology} There is a difference to be made between
community-topic, word topic and user roles. Community topic is the 
$\pi$ vector that we get from
the model (figure~\ref{fig:finalThreadAggregationModel}) that decides the 
membership proportion of a user in
different latent communities. Word topic is the $\beta$ vector of word topic
proportions from the LDA component of the model,
figure~\ref{fig:finalThreadAggregationModel}. There is a one to one
correspondence between $\beta$ and $\pi$ vectors as seen in
figure~\ref{fig:finalThreadAggregationModel}. $\beta$ helps us in identifying
what contents users are generally interested in in a given latent community.
 A user role is a specific configuration of $\pi$. It can be just the case
that a role '$r$' might be the $\pi$ vector where $r$-th coordinate is 1
and all else are 0 out of the total K coordinates, i.e. it predominantly relates
to that $r$-th latent community.

\section{Scalable Estimation}
\label{estimation}

\begin{figure}
\begin{center}
\includegraphics[height=6cm,width=9cm]{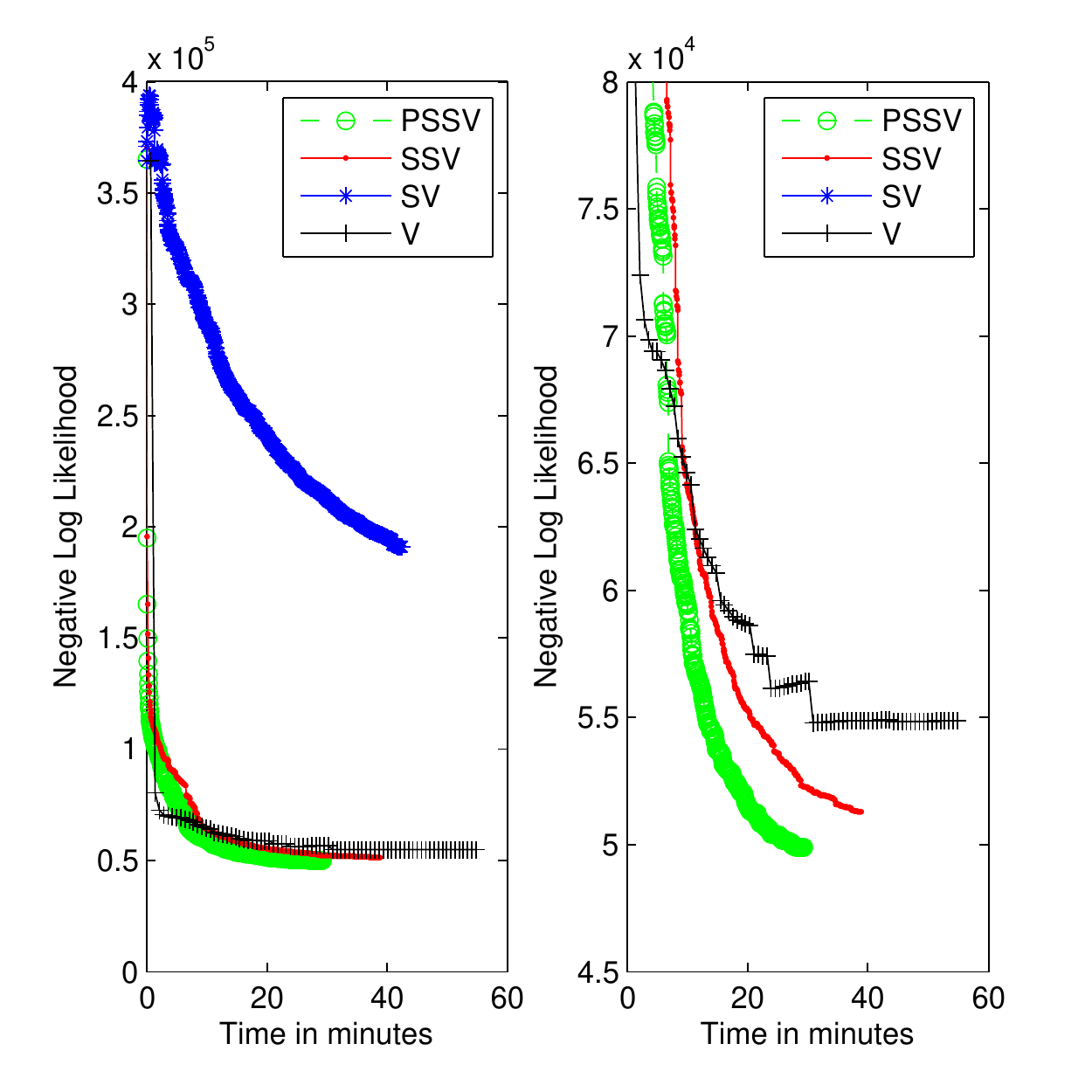}
\end{center}
\caption{The log-likelihood vs incremental speed optimization routine. 
The right hand plot is a zoomed in version of the left. PSSV
(Parallel Sub-sampled Stochastic Variational), SSV(Sub-sampled Stochastic
Variational), SV(Stochastic Variational) and V(Variational). Each addition of
optimization increases the speed by several orders of magnitude. The final PSSV
is 4 times faster than (V)ariational and achieves better log-likelihood too.}
\label{fig:SpeedOptimization}
\end{figure}
The global update equations in previous sections are computationally
very expensive and slow as we need to sum over all the updated local
variables. $U$ users with $T$ threads and vocabulary size $V$ leads to
$O(U^2T+UVT)$ local variables. Traditional sampling or 
variational estimation techniques would be
quite slow for such a model. In order to obtain faster convergence we
make use of stochastic variational approximation along with
sub-sampling and parallelization. 

The updates in case of SVI with sub-sampling follow a two step procedure. Step
one computes a local update for the global variables based on the sub-sampled
updated local variables. The local updates ($\gamma^{'},
\nu^{'}, \lambda^{'}$ and $\tau^{'}$) for the global variables ($\gamma,
\nu, \lambda$ and $\tau$) are

\begin{align}
\gamma_{p,k}^{'} &= \alpha_{k} + \frac{NT}{2|S_p|}\sum_{q \in S_p} \sum_{h}
\! \phi_{t,p,q,k,h} + \frac{NT}{2|S_p|}\sum_{q \in S_p} \! \sum_{g} \!
\phi_{t,q,p,g,k}. 
\label{eqn:gammaUpStoc}
\end{align}

\begin{align}
\nu_{g,h}^{'} &= \nu_{g,h}^{t}+\rho_\nu \frac{NT}{2|S_p|}\sum_{q \in
S_p}\frac{dL}{\partial\nu_{g,h}}.
\label{eqn:nuUpStoc}
\end{align}

\begin{align}
\lambda_{g,h}^{'} &= \frac{\bigg( \sum_{t} \! \sum_{p,q} \! \phi_{t,p,q,g,h}
y_{t,p,q} + \kappa_{g,h} \bigg) }{
 \bigg( \bigg( \sum_{t} \! \sum_{p,q} \! \phi_{t,p,q,g,h} \bigg) + 
\frac{1}{\theta_{g,h}} \bigg) \nu_{g,h}}.
\label{eqn:lambdaUpStoc}
\end{align}

\begin{align}
\tau_{p,v}^{'} = \nu_{v} + \frac{NT}{2|S_p|} 
\bigg(\sum_{w_{t,p,i}=v}^{N_{t,p}} \chi_{t,p,i,k} \bigg). 
\label{eqn:tauUpStoc}
\end{align} 

\IncMargin{1em}
\begin{algorithm}[t]
\small
\SetAlgoLined
Input : $Y,W,P,\alpha,\theta,\kappa,\eta$\\
Initialize : $\gamma\leftarrow \gamma_0$,
$\tau\leftarrow \tau_0, \nu\leftarrow \nu_0, \lambda\leftarrow \lambda_0$\\
\While{not converged}{
\For{$c$ processors \textbf{in parallel}}{
	pick a set of threads $T$
	\For{each $t\in T$}{ 	
		pick a node $p$, $\forall q\in~neighborhood~\delta_{t,p}$\\
		\While{$\phi~\&~\chi$ not converged}{
			get new $\phi_{t,p\rightarrow q}$, $\phi_{t,p\leftarrow q}$,
			$\phi_{t,q\rightarrow p}$, $\phi_{t,q\leftarrow p}$\\
			and $\chi_{t,p,i} \forall i \in N_{t,p}$ \\
			iterate between $\phi$ and $\chi$ using
			equations~\ref{eqn:phiUp}~and~\ref{eqn:chiUp}			
			}
		}
	}
aggregate $\phi$ and $\chi$ obtained from different processors.\\
get local update $\gamma^{'}$,$\tau^{'}$, $\nu^{'}$, $\lambda^{'}$ via
stochastic approximation of
equations~\ref{eqn:gammaUp},\ref{eqn:tauUp},\ref{eqn:nuUp},\ref{eqn:lambdaUp}.\\
get global updates of $\gamma$,$\tau$, $\nu$, $\lambda$; e.g. $\gamma^{t+1} =
(1-step)\gamma^{t}+(step)\gamma^{'}$ \\
Similarly globally update $\tau,\nu,\lambda$ as above using
equation~\ref{eqn:globalUpStoc}.
}
\label{algo:stochasticAlgo}
\caption{PSSV: Parallel Sub-sampling based Stochastic Variational inference for
the proposed model}
\end{algorithm}

where $S_p$ is a set of neighborhood edges of user $p$, and $N$ and $T$ are
total number of edges and threads respectively in the network. The set $S_p$ is
chosen amongst the neighbors of $p$ by sampling equal no. zero and non-zero
edges. 

In step two of the sub-sampled SVI the final update of global variable is
computed by the weighted average of the local updates of the global variable and
the variables value in the previous iteration:

\begin{align}
\mu^{t+1} = (1-\xi^t)\mu^{t} + \xi^t\mu^{'}. 
\label{eqn:globalUpStoc}
\end{align} 

where $\mu$ represents any global variable from $\lambda, \nu, \gamma, \tau$.
 $\xi^t$ is chosen appropriately using SGD literature and is
decreasing.  $\xi^t$ is standard stochastic
gradient descent rate at iteration $t$, also expressed as $\xi^t =
\frac{1}{(t+\zeta)^{\rho}}$~\cite{conf/nips/GopalanMGFB12}. $\zeta$ and
$\rho$ are set as 1024 and 0.5 respectively for all our experiments in the
paper, and $t$ is the iteration number. 

We achieve further speed by parallelizing the text ($\chi$) and network
($\phi$) local variational updates. This is achievable  as the
dependency between $phi$ and $\chi$ parameters (defined in
equations~\ref{eqn:chiDelta} and \ref{eqn:phiDelta}) allows us to parallelize their
variational updates.
Algorithm~\ref{algo:stochasticAlgo} describes the parallelized SVI with
sub-sampling updates for the local parameters.
Figure~\ref{fig:SpeedOptimization} 
shows a plot of how the final (p)arallel (s)ub-sampling based (s)tochastic
(v)ariational (PSSV) inference is faster than each of its individual components.
SO dataset described in section~\ref{sec:dataset} is used as the data for this 
experiment.
The number of parallel cores used in the PSSV scheme is four whereas its one
for the rest of the three.
The amount of sub-sampled forum threads is 400 and the total number of threads is 14,416.
All the schemes in the graph start with the same initialization values of the
hyper-parameters. PSSV is atleast twice as fast as the nearest scheme besides
obtaining the best log-likelihood of all the four at the point of convergence.
The SV (stochastic variational) samples one thread at a time and therefore
takes some time in the beginning to start minimizing the objective value
(negative log likelihood).
The objective value increases in the first few iterations for SV. The number of
iterations to be done by SV is very large but each iterations takes the smallest time of
all four. The V (variational) scheme takes the least number of iterations to
converge though its iterations are the most time consuming as it has to go
through all the 14,416 threads in every iteration.

\begin{table}
\begin{center}
\begin{tabular}{c|c|c|c|c|}
 & users & threads & posts & edges\\\hline
 TS & 22,095 & 14,416 & 1,109,125 & 287,808\\\hline
 UM & 15,111 & 12,440 & 381,199 & 177,336\\\hline
 SO & 1,135,996 & 4,552,367 & 9,230,127 & 9,185,650\\\hline
\end{tabular}
\label{tab:dataStats}
\end{center}
\caption{Dataset statistics. SO mostly has edges with weight one.}
\end{table}

\paragraph{System details} The machine used in all the experiments in this paper
is ``Intel(R) Xeon(R) CPU E5-2450 0 @ 2.10GHz'' 16 corewith 8GBs of
RAM per core. The operating system is Linux 2.6.32 x86\_64. 

\section{Datasets}
\label{sec:dataset}

\begin{figure}
\begin{center}
\includegraphics[height=6cm,width=9cm]{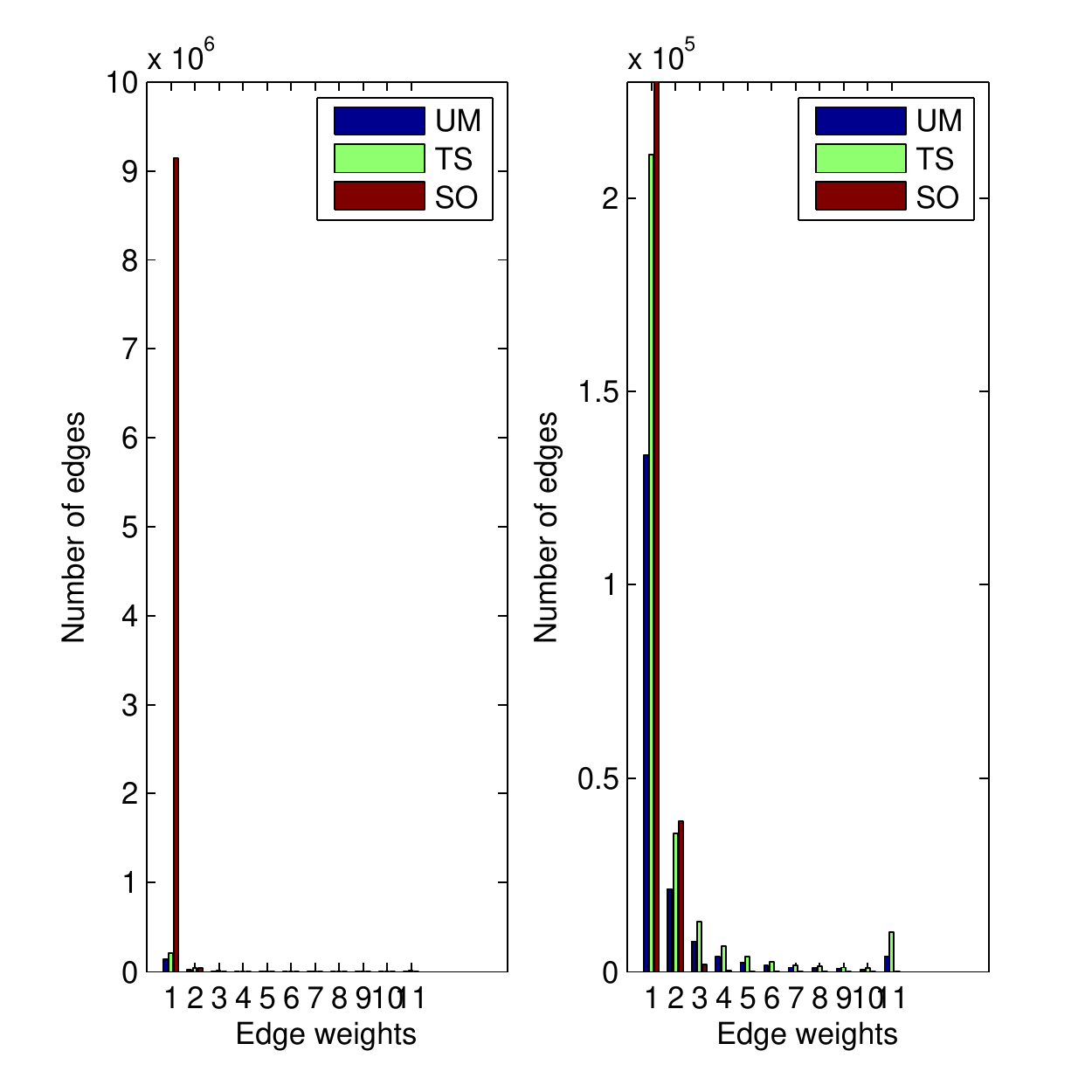}
\end{center}
\caption{Distribution of different edge weights over the 3 datasets. SO 
predominantly consists of edges with weight one. Right hand plot is a scaled
version of the left. The label '11' contains edge weights of 11 and above.}
\label{fig:EdgeDistribution}
\end{figure}

We analyse three real world datasets corresponding to two different forums: 1)
Cancer-ThreadStarter, 2) Cancer-UserName, and 3) Stack Overflow. To test the
stability of the model we use a synthetically generated dataset. 
The Cancer forum \footnote{\url{http://community.breastcancer.org}} 
is a self-help community where users who either have cancer, are 
concerned they may have cancer, or care for others who have cancer, 
come to discuss their concerns and get advice and support.
StackOverflow is an
online forum for question answering primarily related to computer science. We
use the latest dump of Stack
Overflow~\footnote{\url{http://www.clearbits.net/torrents/2141-jun-2013}}. In
each of these datasets a user posts multiple times in a thread and all these
posts are aggregated into one bigger posts per thread as defined in
section~\ref{sec:gen-story}.
Number of times a user $u$ replies to user $v$ in thread $T$ is the edge weight
of edge $u\rightarrow v$ in thread $T$.
Table~\ref{tab:dataStats} gives the distributions of edges, posts, users and
threads in the three datasets used. 

\subsection{Cancer-ThreadStarter (TS)}
In the Cancer forum, the conversations happen
in a structured way where users post their responses on a thread by thread
basis. Every thread has a thread starter that posts the first message and
starts the conversation. 
We construct a graph from each thread by drawing a link from each participant on 
the thread to the participant who started the thread
This graph has 22,095 users and 14,416 Threads. 
\subsection{Cancer-Username Mention (UM)}
Users call each other by their usernames (or handle assigned to them in the
forum) while posting in many cases. We create a graph where in an edge between
user $u$ and user $v$ in thread $t$ means that user $u$ calls user $v$ by
username in thread $t$. This graph has 15,111 users and
12,440 threads.

\subsection{Stack Overflow (SO)}
In Stack Overflow users ask questions and then other users reply with their
answers. We obtain the ThreadStarter graph from this structure. This dataset has
have 1,135,996 users and 4,552,367 threads.
\subsection{Synthetic data}
We generate a synthetic dataset using the generative process defined in
section~\ref{sec:gen-story}. We have 1000 users and 100 threads. The number of 
posts and edges vary depending on the choice of priors $\alpha$ and $\eta$

\begin{figure}
\begin{center}
\includegraphics[height=6cm,width=9cm]{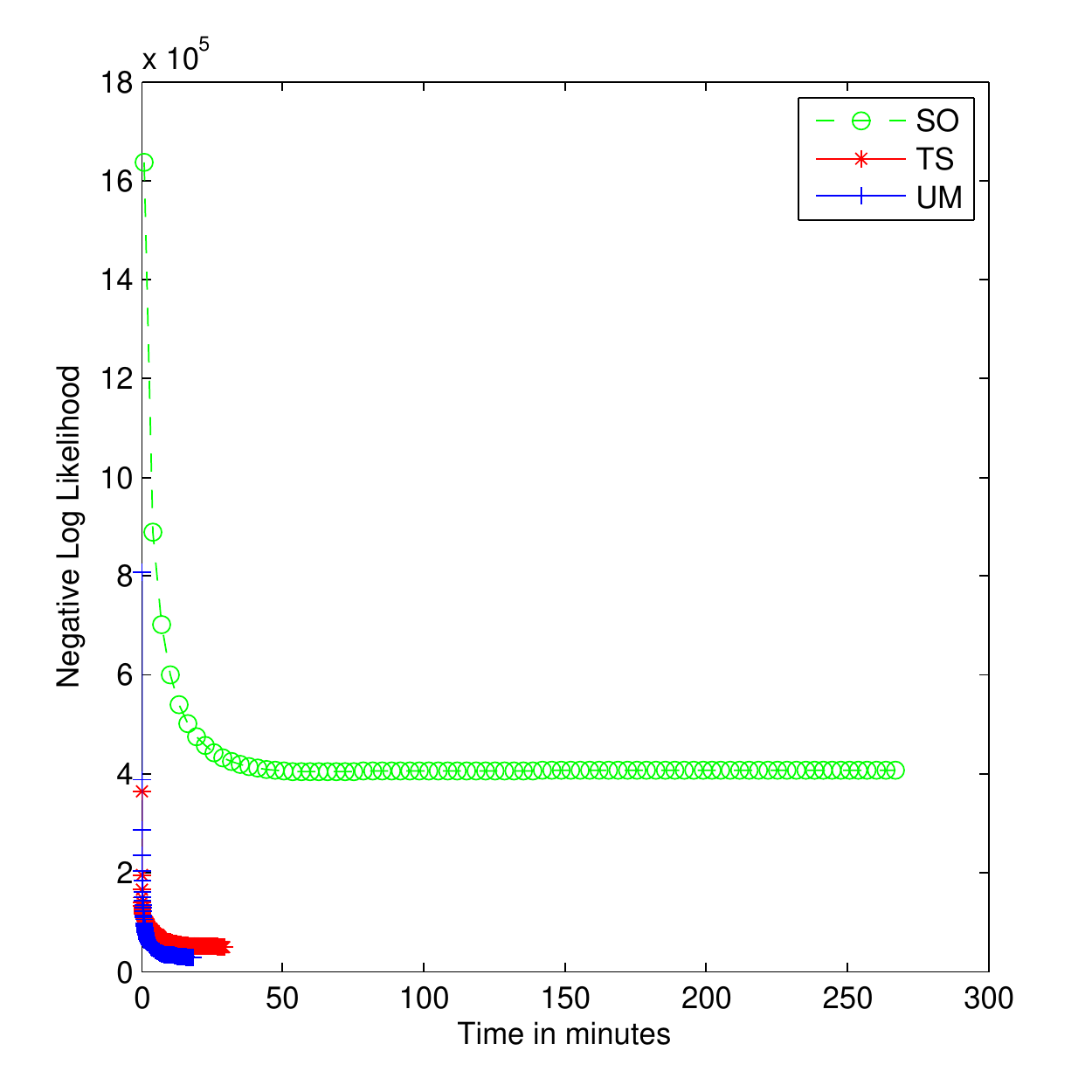}
\end{center}
\caption{The log-likelihood over heldout set for the fully tuned model on the
3 datsets}
\label{fig:finalLLheld}
\end{figure}

\begin{table}
\begin{center}
\begin{tabular}{c|c|c|c|c|c|c|}
 & $\alpha$ & $\omega$ & $\theta$ & $\kappa$ & $\eta$ & $K$\\\hline
 TS & 0.05 & 1e-4 & 2.5$\sim$1.5 & 2.5$\sim$1.5& 0.05 & 10\\\hline
 UM & 0.05 & 1e-3 & 2.0$\sim$1.0 & 2.0$\sim$1.0 & 0.05 & 10\\\hline
 SO & 0.05 & 1e-2 & 1.0$\sim$0.5 & 1.0$\sim$0.5 & 0.05 & 20\\\hline
\end{tabular}
\label{tab:tunedParameters}
\end{center}
\caption{Tuned values for the parameters. $\theta$ and $\kappa$ are matrices
and $a\sim b$ assigned to them means diagonal values are $a$ and non-diagonals
are $b$. $K$ is the number of topics}
\end{table}

\section{Experimental Setup and Evaluation}
\label{sec:setup}
We divide each dataset into three subsets: 1) the training set, 2) the heldout set,
and 3) the test set. We learn our model on training set and tune our priors 
($\alpha, \eta, \kappa, \theta$ etc.) on heldout set. The split is done over the
edges where 80\% of the edges are in training and rest 20\% are divided amongst
heldout and test equally. 
For the two cancer datasets we only predict non-zero edge weights whereas for
the Stack Overflow we predict zero as well as non-zero 
edge weights.
Graph \ref{fig:EdgeDistribution} shows the distribution of edge weights in
cancer and Stack Overflow dataset. We chose Stack Overflow to predict
zero weights since it has large number of edges with very low weights,
predominantly weight one. Predicting zero as well as non-zero edge weights
demonstrates that the model is versatile and can predict a wide range of 
edge-weights. In addition to 20\% of the total
non-zero edges we randomly sample equal number of zero edges from the graph for
the SO held and test set.
The optimization objective for learning is defined in
equation~\ref{eqn:VarLowerBound}.

A link prediction task is incorporated to demonstrate the model's effectiveness.
It is a standard task in the area of graph clustering and social networks in
particular. Researchers have
used it in the past to demonstrate their model's learning 
ability~\cite{Nallapati:2008:JLT:1401890.1401957}.   
The link prediction task works as an important validation of our model. If the
proposed model performs better than its individual parts then it can be safely
concluded that it extracts important patterns from each of its building
blocks. Moreover it adds validity to the qualitative analysis of the results. 

\paragraph{Link-prediction} 
  
 We predict the edge-weight of the edges
present in the test set. The predicted edge, $\hat{Y}_{t,u,v}$, between users $u$ and
$v$ in thread $t$ is  defined as 
\begin{align}
\hat{Y}_{t,u,v} = \pi^T_uB\pi_v\label{eqn:prediction}.\\
B=\nu.*\lambda\label{eqn:blockMat}
\end{align}
and the prediction error is the $rmse$, defined as given predicted edge
$\hat{Y}_{t,u,v}$ and the actual edge $Y_{t,u,v}$,
\begin{equation}
rmse=\sqrt{\sum(\hat{Y}_{t,u,v}-Y_{t,u,v})^2}.
\end{equation}

\begin{figure}
\begin{center}
\includegraphics[height=6cm,width=9cm]{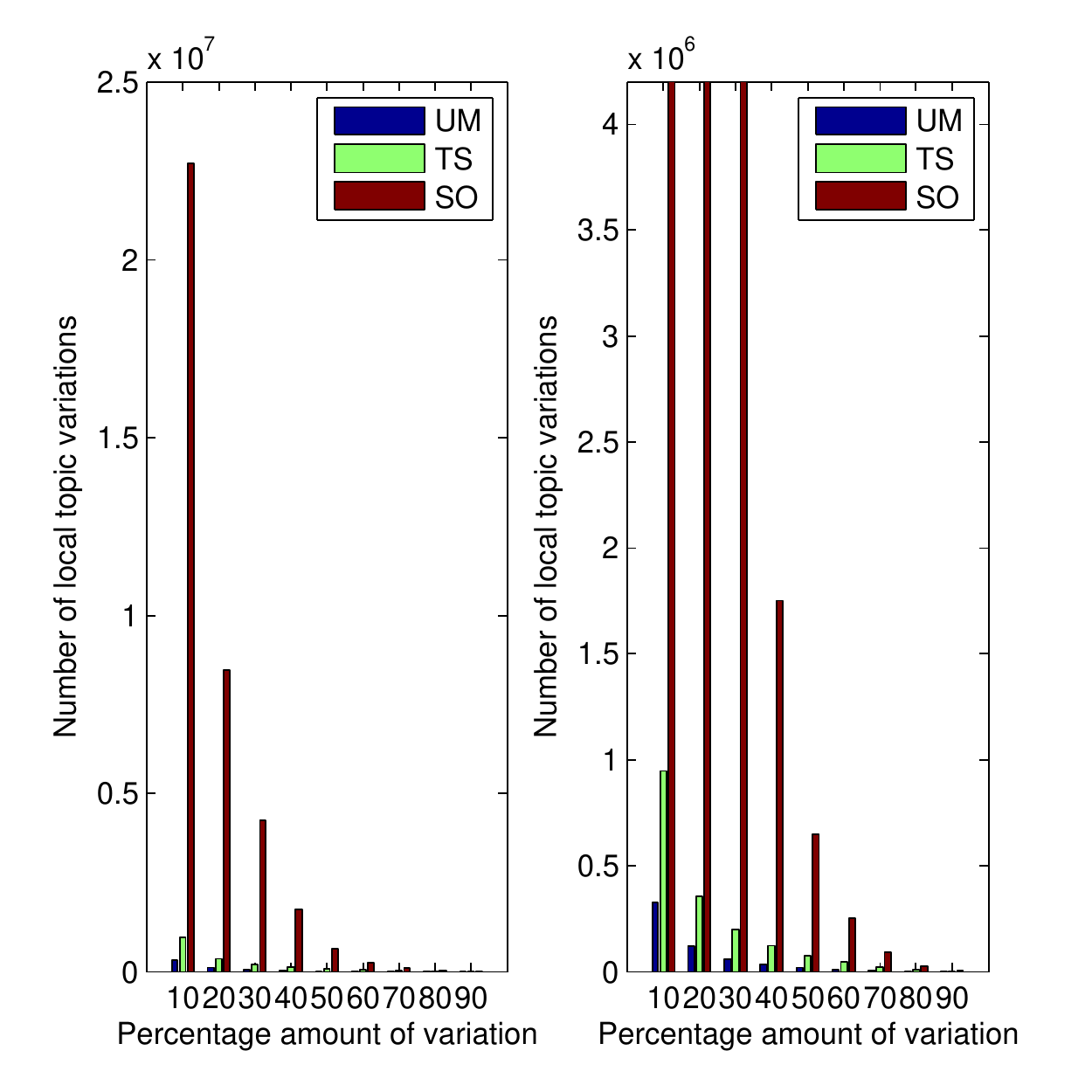}
\end{center}
\caption{Number of local variations in topic proportion on a per user per thread
level. The axis is percentage variation (from 10 to 90 percent). The right hand
plot is a scaled in version of the left.}
\label{fig:localTopicVariations}
\end{figure}

\begin{center}
\begin{figure*}
\includegraphics[height=7cm,width=18cm]{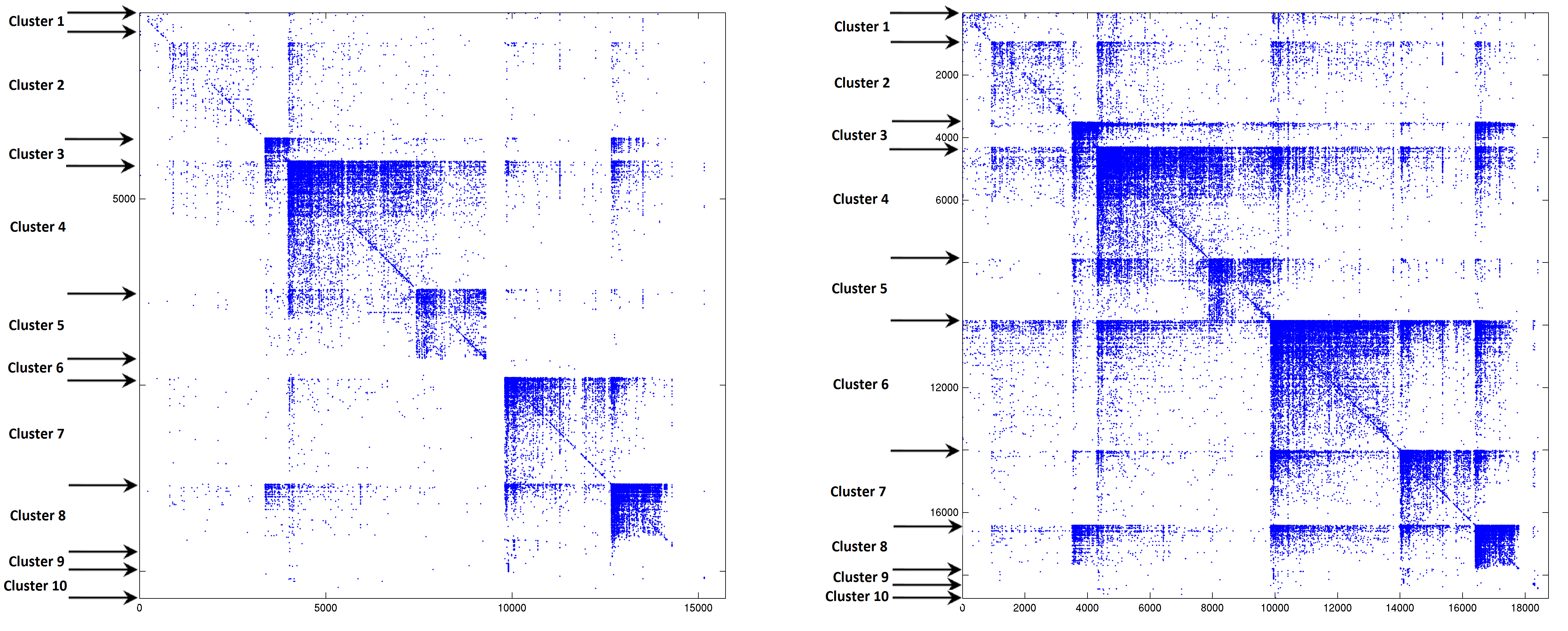}
\caption{Adjacency matrix of users sorted by clusters. Left side is 
clustered by MMSB and right side is clustered by our model using user's dominant
role as cluster index over TS dataset. Our model is able to correctly cluster 3K
additional users that MMSB doesnt assign any dominant cluster (or role) and
discovers a new role (Cluster-6).}
\label{fig:similarityMatTS}
\end{figure*}
\end{center}
The summation is over the edges in the test (or heldout) set. The block matrix
$B$ described in equation~\ref{eqn:blockMat} is well defined for both MMSB and
the proposed model. Hence the prediction is obtained for the active network modelling without
LDA (just MMSB component) and with LDA. We created an artificial
weighted Identity matrix for LDA $\hat{B}=m*I$. It is a diagonal matrix with
all element values $m$. For every user $u$ and every thread $t$ the topics discovered over the
posts of $u$ in $t$ is used as the vector $\pi_u$ in
equation~\ref{eqn:prediction} for prediction. A diagonal $B$ is desirable in
block models as it provides clean separation among the clusters
obtained~\cite{Airoldi:2008:MMS:1390681.1442798}.
The value of $m$ is tuned over heldout set. We define a basic baseline that
always predicts the average weight ($\bar{Y}$) all the edges, zero (Stack
Overflow) or non-zero (Cancer), in the heldout or test set.
\begin{equation}
rmse_{baseline} = \sqrt{\sum(\bar{Y}-Y_{t,u,v})^2}.
\end{equation}

\paragraph{Parameter tuning}
We tune our parameters $\eta, \kappa,
\theta, \alpha$, and $K$ (number of community-topics) over the held set.
$\omega$, the parameter to balance the contribution of the text side to the 
network side is tuned over the
heldout set. It is used in the local variational update of
$\phi$(equation~\ref{eqn:phiDelta}).
Equation~\ref{eqn:phiDelta} contains a summation term over all the tokens
$\sum_{i=1}^{N_{T_p}}\chi_i$ in the per user per thread document and if  not
balanced by $\omega$ will dominate the rest of the terms. The constant $\epsilon$ used in
equation~\ref{eqn:phiDelta} and~\ref{eqn:chiDelta} is a smoothing constant and
is fixed at a low value. The six quantities, $\alpha, \omega, \theta, kappa,
\eta$ and $K$ are tuned in that sequence. $\alpha$ is tuned first keeping rest
constant then $\omega$ and so on where each next to be tuned parameter uses
values of already tuned parameters. Table~\ref{tab:tunedParameters}
shows the final values of all the parameters.

Figure~\ref{fig:finalLLheld} shows plot of tuned log-likelihood over the 3
datasets against time. UM being the smallest of the two takes
the least amount of time. 

\section{Results}
\paragraph{Link prediction} Table~\ref{tab:predictionResults} shows the link
prediction results on heldout and test set for the for the four prediction
model.
\begin{table}
\begin{center} 
\begin{tabular}{p{1cm}|p{0.7cm}|p{0.7cm}|p{0.7cm}|p{0.7cm}|p{0.7cm}|p{0.7cm}|}
  & UM held & UM test & TS held & TS test & SO held & SO test \\\hline
Our Model & \textbf{1.303} &\textbf{1.292} &\textbf{2.782} & \textbf{2.748} & 
\textbf{0.348}& \textbf{0.361} \\\hline 
MMSB &1.450& 1.502 & 2.984& 2.881 &0.421 & 0.434 \\\hline 
LDA &1.793& 1.731	&3.725 & 3.762 &0.466 & 0.479\\\hline
Baseline &1.886& 1.983 &4.504 &4.417&0.502& 0.509\\\hline
\end{tabular}
\caption{Link prediction results over the 3 datsets }
\label{tab:predictionResults}
\end{center}
\end{table}

The proposed approach to model thread level conversational roles outperforms
all of the other models. LDA performs poorer than MMSB since LDA does not 
explicitly model network information. 
\paragraph{Cancer dataset}

Figure~\ref{fig:localTopicVariations} shows the number of times the global role
of a user is different from the thread level role that he plays. It is 
interesting to see that the variation between global and thread level role
assignment is high among all the datasets. A model that ignores this local vs
global dynamics tends to lose a lot of information.
Figure~\ref{fig:similarityMatTS} shows the plot of the user by user adjacency 
matrix for TS dataset. The users are sorted based on the community-topic cluster
(roles) assigned by the respective models (our model and MMSB model). The number of
community-topics are 10 and every user is assigned the dominant
community-topic, $\pi$, (role) that they have more than 50\% of chance of lying in. A user 
is discarded if he
doesn't have the said dominant role. Our model is unable to assign a clear 
role to 3.3K users and
the MMSB approximately to 6.3K users out of 22K. Based on the topics assigned,
users are sorted and their adjacency matrix shows clean clustering along the
block diagonals.
As seen in the figure, the combined model is able to effectively find the
primary roles (dominant topic) for the extra 3K users that the MMSB model was
unable to provide for. Besides a new role (Role 6) that is not accounted for by
MMSB is discovered by the proposed model (figure~\ref{fig:similarityMatTS}).
Users that predominantly have role 6 on a global scale tend to vary their roles
often on a thread level, i.e. their topic probabilities change quite often. The
average change in topic probabilities per role per user-thread pair across the
10 roles discovered in TS is 30.6\%; for role 6 it is 41.5\% (highest of all the roles). 
This means
that this role is very dynamic and an active sub-network modelling helps here as 
it captures the
dynamism of this entity. From figure~\ref{fig:similarityMatTS} cluster of roles
4, 5, 6, 7 and 8 are the largest. Table~\ref{tab:top15WordsTS} corresponds to
top 15 words corresponding to these roles. Role 4, role 7 and role 8 are
related to discussion regarding cancer where as role 5 is related to
conversations regarding spiritual and family matters.  But role 6 does not seem
to be related to any specific type of conversation. It is free flowing and has
lots of non specific words which tells us that there is a cornucopia of
discussions happening in this role with no specific matter at hand. This fact
is also verified by looking at the raw Cancer forum data. Users who are
predominantly in this role tend to post across many discussion threads and
variety of conversation topics. This role is detected by our model due to the
fact that it takes into account the dynamics of such a role.

\begin{figure*}
\centering
  \includegraphics[width=1\linewidth]{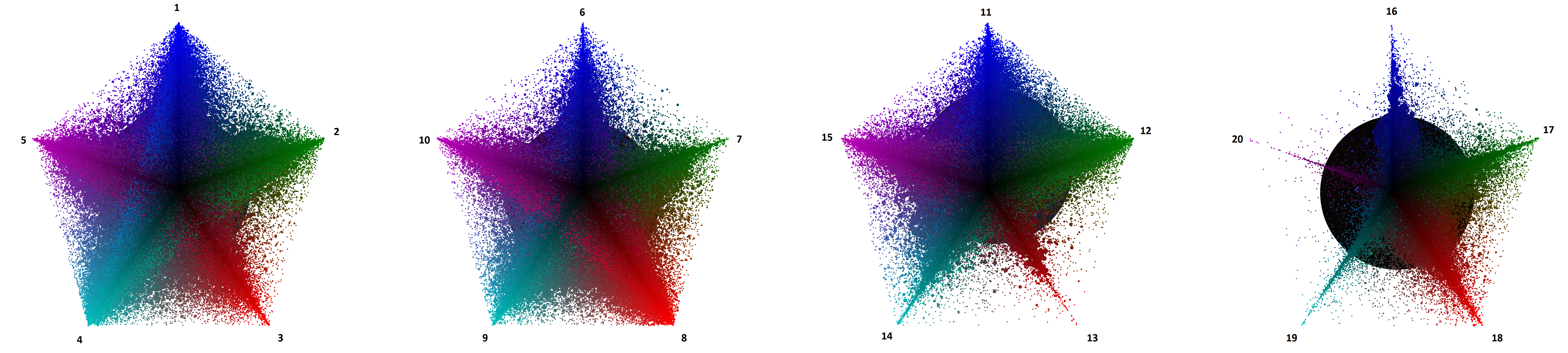}
\caption{The 20 roles assigned to users in the stack overflow
dataset. The numbers at the vertex are the role numbers. Due to the large
number of roles we visualize them 5 at a time with first 5 first then second
5 and so on.
We can see that the roles are separated cleanly and clustered around the
pentagon corners.}
\label{fig:SOclusters}
\end{figure*}

\begin{figure}
\centering
  \includegraphics[width=0.7\linewidth]{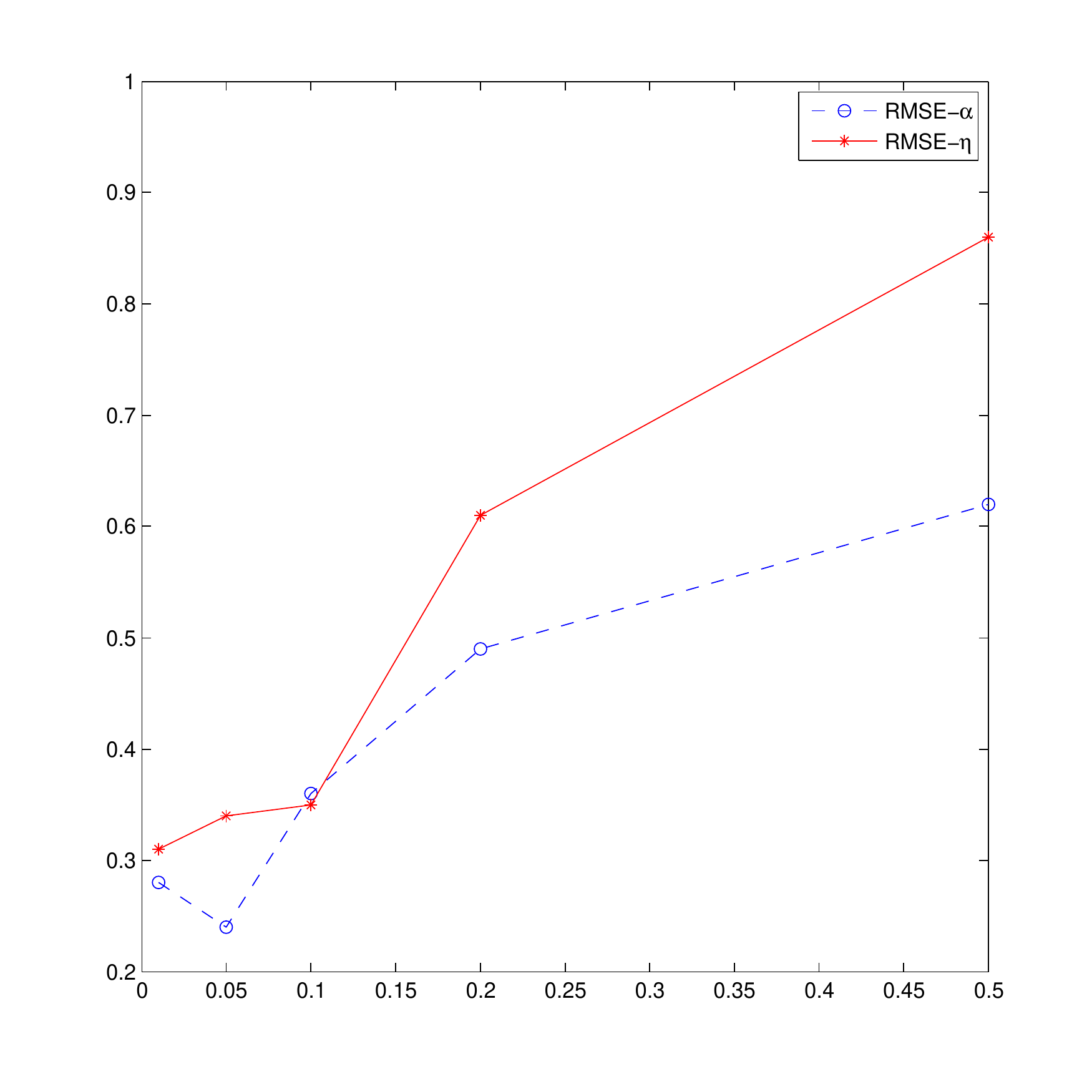}
\caption{RMSE vs $\alpha$ and $\eta$ for the synthetic dataset. The X-axis is
$\alpha$ or $\eta$ values and the Y-axis is the RMSE of the recovered $\pi$. }
\label{fig:syntheticRMSE}
\end{figure}

\begin{table}
\begin{center} 
\begin{tabular}{c|c|c|c|c}
Topic 4  & Topic 5 & Topic 6 & Topic 7 & Topic 8 \\\hline
side            &same   &their  &surgeon        &radiat \\\hline
test            &life   &live   &everi  &anoth\\\hline
took            &tell   &happi  &found  &doctor\\\hline
away            &mani   &mayb   &down   &problem\\\hline
left            &famili &sorri  &alway  &pleas\\\hline
support         &prayer &best   &while  &person\\\hline
doesn           &though &check  &home   &kind\\\hline
seem            &ladi   &news   &bodi   &soon\\\hline
move            &until  &question       &these  &each\\\hline
almost          &wish   &dure   &bone   &hard\\\hline
scan            &someon &deal   &mean   &might\\\hline
idea            &under  &case   &came   &medic\\\hline
studi           &felt   &mind   &posit  &herceptin\\\hline
guess           &where  &seem   &drug   &share\\\hline
diseas          &nurs   &haven  &send   &free\\\hline

\end{tabular}
\caption{top 15 words for topics corresponding to top 5 biggest role in
TS.}
\label{tab:top15WordsTS}
\end{center}
\end{table}



\paragraph{Stack Overflow} 
The optimal topic number for SO dataset is 20 community-topics as noted in
table~\ref{tab:tunedParameters} and the number of users are 1.13 million. It is
difficult to visualize the user-user adjacency matrix of this size. The 20
topic set is divided into four sets with size 5 each. Topics 1 to 5
form set one, topics 6 to 10 form set two and so on. Every user's role is
visualized by projecting user's $\pi$ over a pentagon as shown in
figure~\ref{fig:SOclusters}. The projection uses both position and
color to show values of community-topic $\pi$ for each user.
Every user $u$ is displayed as a circle (vertex) $v_u$ in the figure where the
size of the circle is the node degree of $v_u$ and position of $v_u$ is equal to
a convex combination of the five pentagon corner coordinates $(x, y)$ that are weighted by the
elements of $\pi_u$. Hence circles $v_u$ at the pentagon's corners
represent $\pi$'s that have a dominating community  in the 5
community-topics chosen, while circles on the lines connecting the corners 
represent $\pi$'s with
mixed-membership in at least 2 communities (as only a partial $\pi$ vector is
used in each sub-graph).
All other circles represent $\pi$'s with mixed-membership in $\geq 3$
communities.
Each circle $v_u$'s color is also a $\pi$-weighted convex
combination of the RGB values of 5 colors: blue, green, red, cyan and purple. This
color coding helps distinguish between vertices with 2 versus 3 or more
communities. We observe a big black circle at the back ground of every plot.
This circle represents the user with id 22656
\footnote{\url{http://stackoverflow.com/users/22656/jon-skeet}} that has the highest node-degree of 25,220 in the SO dataset. This user has the highest all time reputation on
stack overflow and tends to take part in myriads of question answering threads.
Hence he is rightly picked up by the model to be in the middle of all the roles. 

Figure~\ref{fig:SOclusters} has a clean clustering where the nodes are clustered
around the pentagon vertices. This indicates that the model is able to find
primary roles for most of the users in the forum.  Though
table~\ref{fig:localTopicVariations} shows that there is significant amount of
variation with respect to the global role of a user at a thread level. 
Modeling this variation helps our model in
getting better clusters as compared to simple MMSB; this fact is apparent from
the link prediction task too, table~\ref{tab:predictionResults}. We get an rmse
of 0.348 on heldout and 0.361 on test set which is better than all the other 3
approaches. 

Table~\ref{tab:top15WordsSO} shows the top 15 words corresponding to first 6
roles discovered in SO. While roles 1, 2, 3 are pertinent to discussions
regarding database conversations, general coding, android
and J2EE, role 4 relates to online blogs and browsing apis. Role 5 and role 6
are related to server/client side browser based coding and general coding
respectively.

\begin{table}
\begin{center} 
\begin{tabular}{c|c|c|c|c|c}
Topic 1  & Topic 2 & Topic 3 & Topic 4 & Topic 5 & Topic 6 \\\hline
public  &code   &function       &that   &name   &array \\\hline
valu    &should &also   &view   &method &more \\\hline
chang   &your   &then   &thread &time   &what \\\hline
when    &user   &object &into   &document       &system \\\hline
event   &properti       &creat  &control        &line   &about \\\hline
would   &blockquot      &follow &current        &element        &sure \\\hline
list    &just   &form   &link   &differ &post \\\hline
databas &field  &implement      &oper   &would  &each \\\hline
there   &class  &valu   &question       &defin  &imag \\\hline
issu    &overrid        &server &length &file   &class \\\hline
path    &main   &veri   &creat  &result &paramet \\\hline
display &result &string &each   &where  &applic \\\hline
like    &size   &start  &result &more   &just \\\hline
order   &import &java   &blog   &know   &local \\\hline
save    &project        &android        &featur &browser        &specif \\\hline
\end{tabular}
\caption{top 15 words for word topics corresponding to first 6 role clusters in
SO.}
\label{tab:top15WordsSO}
\end{center}
\end{table}
Do demonstrate the effectiveness of our model on this dataset, we take the
example of user 20860. User id 20860 is globally assigned role 1 as the dominant
role but he also takes part in coding related discussions. For example, 
\begin{itemize}
  \item \textit{Because the join() method is in 
the string class, instead of the list class?
I agree it looks funny.}  
\item \textit{The simplest solution is to use
shell\_exec() to run the mysql client with the SQL script as input. 
This might run a little slower because it has to fork, but you can write 
the code in a couple of minutes and then get back to working on something useful. 
Writing a PHP script to run any SQL script could take you weeks\ldots.}

\end{itemize}

But in most of the cases he visits general software or coding questions
that are specifically related to databases and this fact is picked up by our
model and it assigns him predominantly (>0.8) a database role even though he 
is active contributor to
general software and coding discussions. MMSB on the other hand assins him 30\%
database (role 1) 30\% general coding (role 2) and rest is distributed among the
remaining 18 roles.

The model picks up other similar cases for which it is able
to successfully distinguish (compared to MMSB) between user's global and local
roles even though they are dynamic in nature.


\paragraph{Synthetic dataset}
Figure~\ref{fig:syntheticRMSE} gives the rmse of the model for the recovery of
community topic $\pi$ over the synthetic dataset. From the graph, higher values
of the parameters make it harder to recover the $pi$ values. For this experiment
we fix topic number at 5 and vary $\alpha$ or $\eta$ by fixing the other at
0.01. The other priors such as $\kappa, \theta,
\omega$ etc. are fixed at the values used to generate the dataset.  
It is apparent that the rmse is more sensitive towards $\alpha$ values and
recovers them well compared to $\eta$. The results are averaged over 20 random
runs of the experiment for the given values of $\alpha$ and $\eta$. The rmse
achieved for lower values of priors $\alpha$ and $\eta$ is very promising as
it means that the confidence interval of the estimate is very high for sparse
priors for this model given sufficient data. 

\section{Related Work}

White et al.\cite{ICWSM124638} proposed a mixed-membership model that obtained
membership probabilities for discussion-forum users for each statistic
(in- and out-degrees, initiation rate and reciprocity) in various profiles and
clustered the users into ``extreme profiles'' for user role-identification
and clustering based on roles in online communities,. Ho et al.
\cite{Ho:2012:DHT:2187836.2187936} presented TopicBlock that combines text and
network data for building a taxonomy for a corpus.
The LDA model and MMSB models were combined by
Nallapati et al. \cite{Nallapati:2008:JLT:1401890.1401957} using the
Pairwise-Link-LDA and Link-PLSA-LDA models where documents are assigned
membership probabilities into bins obtained by topic-models. Sachan et
al.~\cite{Sachan:2012:UCI:2187836.2187882} provide a model for community
discovery that combines network edges with hashtags and other heterogeneous data
and use it to discover communities in twitter and Enron email dataset.

For simultaneously modeling topics in bilingual-corpora, Smet et al.
\cite{Smet:2011:KTA:2017863.2017915} proposed the Bi-LDA model that generates
topics from the target languages for paired documents in these very languages.
The end-goal of their approach is to classify any document into one of the
obtained set of topics. For modeling the behavioral aspects of entities and
discovering communities in social networks, several game-theoretic approaches
have been proposed (Chen et al. \cite{Chen:2010:GFI:1842547.1842566}, Yadati and
Narayanam \cite{Yadati:2011:GTM:1963192.1963316}). Zhu et
al.~\cite{Zhu:getoor:MMSB-text} combine MMSB and text for link prediction and
scale it to 44K links.

Ho et al.~\cite{HoYX12} provide  unique triangulated sampling schemes for scaling
mixed membership stochastic block models~\cite{Airoldi:2008:MMS:1390681.1442798} to
the order of hundreds of thousands of users. Prem et
al.~\cite{conf/nips/GopalanMGFB12} use stochastic variational inference 
coupled with sub-sampling techniques to
scale MMSB like models to hundreds of thousands of users.

None of the works above address the sub-network dynamics of thread based
discussion in online forums. Our work is unique in this context and tries to
bring user role modelling in online social networks closer to the
ground realities of online forum interactions.
Active sub-network modelling has been used recently to model gene interaction 
networks~\cite{Lichtenstein:Charleston}. They
combine gene expression data with network topology to provide bio-molecular 
sub-networks, though their approach is not scalable as they use simple EM for
their inference. We leverage the scalable aspects of
SVI~\cite{Hoffman:2013:SVI} to combine MMSB (network topology) with LDA (post
contents) in a specific graphical structure (thread structure in the forum) to
obtain a highly scalable active sub-network discovery model.

Matrix factorization and spectral learning based approaches are some of the
other popular schemes for modelling user networks and content. In recent
past both approaches have been made scalable to a million order node size graph
~ \cite{Gemulla:2011:LMF,Dhillon:2005:FKM}. But these methods are unable to incorporate 
the rich structure that a probabilistic modelling based method can take into account.
For example modelling active sub-networks besides incorporating content as well
as network graph will be very hard to achieve in matrix factorization or 
spectral clustering paradigm.

\section{Discussion and Future Work}
The  proposed model relies on the fact that forum users have dynamic role
assignments in online discussions and leveraging this fact helps to increase
prediction performance as well as understand the forum activity. The model
performs very well in its prediction tasks. It outperforms all the other methods
over all the datsets by a huge margin. The model is scalable and is able to run
on social network dataset of unprecedented content size. There is no past
research work that scales forum contents to more than one  million user and
around 10 million posts. 

The idea that active subnetwork is useful in modelling online forums is
demonstrated qualitatively and quantitatively. Quantitatively it provides better
prediction performance and qualitatively it captures the dynamics of user
roles in forums. This dynamism can help us find new cluster roles that may have
been missed by state of the art clustering approaches, as we observed for
TS dataset. From the synthetic experiments it is observed that the model
recovers its parameters with high likelihood with sparse priors. This works to
its advantage for scalable learning as big data sets tend to be sparse.

The scalability aspects of the inference scheme proposed here are worth noting.
Besides the multi-core and stochastic sub-sampling components of the proposed
inference, the use of Poisson to model the edge weights has enabled us to 
ignore zero-edges if need be. This reduces the amount of work needed for
learning the network parameters. The learned network is at par with the state of the art
inference schemes as demonstrated in the prediction tasks.   

One aspect to explore in future is to combine multiple types of links in
network. For example in many online forums users explicitly friend other users,
follow other users or are members of same forum related sub-groups as other
users. All these relations can be modelled as a graph. It is worth finding out
how important is modelling active sub-network in such a case. It is
possible that various types of links might reinforce each other in learning
the parameters and thus will obviate the need to model a computationally
costly sub-network aspect. As we saw in figure~\ref{fig:syntheticRMSE} that
sparsity helps, but how sparser can we get before we start getting poor results
needs some exploration.

As we have
seen, figure ~\ref{fig:syntheticRMSE}, that the model recovers the
community-topic parameters with very high likelihood for lower values of model
priors $\alpha$ and $\tau$. If this is a general attribute of active sub-network
models then it can be leveraged for sparse learning. Moreover, although in case
of large online forums modelling active sub-networks is computationally
challenging and costly, the sparsity aspects of active sub-networks might help
reduce the computation costs.

\bibliography{mybib}
\bibliographystyle{plain}
\appendix
\label{sec:appendix}

The log-likelihood of the model:
\begin{align}
\log L &= \log \! P(Y, W, Z_{\leftarrow}, 
Z_{\rightarrow}, \Pi, B, \beta | \alpha, \eta, \theta, \alpha) \nonumber\\
\nonumber &= \sum_{t} \bigg[ \sum_{p,q} \! \log P(Y_{t,p, q} | Z_{t,p \rightarrow q} 
     , Z_{t,p \leftarrow q}, B) \\ \nonumber 
     &+ \sum_{p,q} \log P(Z_{t, p \rightarrow q} | \Pi_q) \\ \nonumber 
     & + \sum_{p,q} \log \! P(Z_{t, p \leftarrow q} | \Pi_{q}) \bigg] 
     + \sum_{p} \log \! P(\Pi_{p} | \alpha)  \\ \nonumber 
     & + \bigg[ \sum_{t=1}^{T} \! \sum_{p \in t} \sum_{i=1}^{N_{T_{p}}} 
     \log \! P(w_{t,p,i} | Z'_{t,p,i}, \beta) \\ \nonumber 
     & + \sum_{t=1}^{T} \sum_{p \in t} \sum_{i=1^{N_{T_{p}}}} \log \! 
     P(Z'_{t,p,i} | \bar{Z}_{t, p \rightarrow q}) \bigg]   
     \\ \nonumber & + \sum_{k} \log P(\beta_{k} | 
     \eta) +  \sum_{g,h} \log P(B_{g,h} | \kappa, \theta).\\ 
     \label{eqn:LL}
\end{align}

The data likelihood for the model in figure~1

\begin{align}
P(Y, R_{p} | \alpha, \beta, \kappa, \eta) &=  \nonumber\\ 
 \int_{\Phi} \!
\int_{\Pi} \sum_{z} \! P(Y, R_{p}, & z_{p \rightarrow q}, z_{p \leftarrow q},
\Phi, \Pi | \alpha, \beta, \kappa, \eta)  \nonumber \\  \nonumber
= \int_{\Phi} \! \int_{\Pi} \sum_{z} \! \bigg[ \prod_{p,q} & \prod_{t}
P(Y_{pq}^{t} | z_{p \rightarrow q}^{t}, z_{p \leftarrow q}^{t}, B) 
\cdot P(z_{p \rightarrow q}^{t} | \Pi_{p}) \nonumber
\\  \cdot P(z_{p \leftarrow q}^{t} |
\Pi_{q})   & \cdot \bigg(\prod_{p} P(\Pi_{p} | \alpha) \prod_{t} \prod_{p}
P(R_{p}^{t} | z_{p \rightarrow q}^{t}, \Phi) \nonumber
\\ \cdot \prod_{k} P(\Phi_{k} |
\beta)&\bigg) \cdot \prod_{g,h}P(B_{gh} | \eta, \kappa) \bigg].
\end{align}

The complete log likelihood of the model is:

\begin{align}
\log \! &P(Y, W, z_{\rightarrow}, z_{\leftarrow}, \Phi, \Pi, B | \kappa, \eta,
\beta, \alpha) \nonumber
\\ & = \sum_{t} \! \sum_{p,q} \! \log P(Y_{pq}^{t} | z_{p
\rightarrow q}^{t} , z_{p \leftarrow q}^{t}, B) \nonumber
\\ &+ \nonumber \sum_{t} \!
\sum_{p,q} \! (\log P(z_{p \rightarrow q}^{t} | \Pi_{p}) + \log \! P(z_{p \leftarrow q}^{t} |
\Pi_{p})) \\  
&+ \sum_{p} \! \log \! P(\Pi_{p} | \alpha) ~+ \sum_{t} \!
\sum_{p} \! \sum_{w \in R_{p}^{t}} \log P(w | z_{p \rightarrow}, \Phi)
\nonumber\\ 
&+ \sum_{k} \! \log P(\Phi_{k} | \beta) + \sum_{gh} \! \log P(B_{gh} | \eta,
\kappa).
\end{align}

The mean field variational approximation for the posterior is 

\begin{align}
q(z, &\Phi, \Pi, B | \Delta_{z_{\rightarrow}}, \Delta_{\Phi}, \Delta_{B},
\Delta_{z_{\leftarrow}}, \Delta_{B_{\kappa}})  = \nonumber \\ \prod_{t} \!
& \prod_{p,q} \! \bigg( q_{1}(z_{p \rightarrow q}^{t} | \Delta_{z_{p \rightarrow
q}}) + q_{1}(z_{p \leftarrow q}^{t} | \Delta_{z_{p \leftarrow q}})  \bigg) \nonumber \\
\cdot \prod_{p} &\! q_{4}(\Pi_{p} | \Delta_{\Pi_{p}}) \prod_{k} q_{3} (\Phi_{k}
| \Delta_{\Phi_{k}}) \prod_{g,h} \! q(B_{g,h} | \Delta_{B_{\eta}},
\Delta_{B_{\kappa}}).
\end{align}

The lower bound for the data log-likelihood from jensen's inequality is: 

\begin{align}
&L_{\Delta} = E_{q}\bigg[ \log \! P(Y, W, z_{\rightarrow}, z_{\leftarrow}, \Phi,
\Pi, B | \kappa, \eta, \beta, \alpha) - \log \! q \bigg]\nonumber\\
&= E_{q} \Bigg[ \sum_{t} \! \sum_{p,q} \! \log \left(
B_{g,h}^{Y_{p,q}^t} \frac{e^{-B_{gh}}}{Y_{pq}^{t}!} \right) +
\sum_{t} \! \sum_{pq} \! \log\left( \prod_{k} (\pi_{p,k}^{z_{p \rightarrow q} =
k}) \right) \nonumber\\
&+ \sum_{t} \! \sum_{p,q} \log \! \left(
\prod_{k}(\pi_{q,k})^{z_{p \leftarrow q} = k} \right)\nonumber\\ 
&+\sum_{t} \! \sum_{p} \! \sum_{w\in R_p^t}  \log \! \left(
\prod_{u\in V}(\bar{z}^T\phi_u)^{w = u} \right)
\nonumber\\ &+ 
\sum_{p} \! \log \left( \prod_{k}
(\Pi_{p,k})^{\alpha_{k} - 1} \cdot \frac{\Gamma(\sum \alpha_{k})}{\prod_{k}
\Gamma(\alpha_{k})} \right) \nonumber\\ & + 
\sum_{k} \! \log\left( \prod_{u\in V}
(\phi_{k,u})^{\beta_{k} - 1} \cdot \frac{\Gamma(\sum \beta_{k})}{\prod_{k}
\Gamma(\beta_{k})} \right) \nonumber\\ &+
 \sum_{g,h} \! \log \! \left( B_{g,h}^{\kappa - 1} /
\eta^{\kappa} \Gamma(\kappa) \cdot \exp(-B_{g,h}/\eta) \right) \Bigg]
\nonumber\\ 
& -E_{q} \Bigg[ \sum_{t} \! \sum_{p,q} \log \big( \prod_{k} (\Delta_{z_{p
\rightarrow q}, k})^{z_{p \rightarrow q}=k} \big) \nonumber \\&+ \sum_{t} \!
\sum_{p,q} \! \log \! \left(
\prod_{k} \! (\Delta_{z_{p \leftarrow q}, k})^{z_{p \leftarrow q} = k} \right)
  \nonumber \\
 &+\sum \! \log \left( \prod_{k} \! (\Pi_{p,k})^{\Delta_{\pi_{pk}}-1}
\frac{\Gamma(\Delta_{\Pi_{p}})}{\prod_{k=1} \! \Gamma(\Delta_{\Pi_{p,k}})}
\right) \nonumber \\ &+ 
\sum_{k} \log \! \left( \prod_{u \in v}
(\Phi_{k,u})^{\Delta_{\Phi_{ku}} - 1)} \frac{\Gamma(\Delta_{\Phi_{k}})}
{\prod_{u \in v} \! \Gamma(\Delta_{\Phi_{k,u}})} \right) \nonumber \\ 
&+ \sum_{g,h} \log \! \left(
\frac{B_{g,h}^{\Delta_{\kappa = 1}}}{\Delta_{\eta}^{\Delta_{\kappa}}
\Gamma(\Delta_{\kappa})} \exp(-B_{g,h}/\Delta_{\eta}) \right) \Bigg].
\label{eqn:VarLowerBound}
\end{align}

$\Delta_{\phi}$ used in the update of $\phi$ in equation~\ref{eqn:phiUp}. The
parameter $\omega$ is used here to balance out the contribution from the text
side to the network. 

\begin{align}
\Delta_{\phi^{'}_{t,p,g,h}} &= y_{t,p,q}( \log \! \lambda_{g,h} + 
\Psi(\nu_{g,h})) - \nu_{g,h} \lambda_{g,h} - \log \! (y_{t,p,q}!)
\nonumber \\ & + \Psi(\gamma_{p,g}) - \Psi(\sum_{g} \gamma_{p,g})
\nonumber \\  & + \Psi(\gamma_{q,h}) - \Psi(\sum_{h} \gamma_{q,h})
\nonumber \\  & + \omega\sum_{i=1}^{N_{T_{P}}} \! \chi_{t,p,i,g} 
\bigg[ \ln \! \frac{\epsilon}{\delta_{p,t}} - \frac{1}{\delta_{t,p}} 
+ \ln \bigg( 1 + \frac{\epsilon}{\delta_{p,t}} \bigg) 
\cdot \frac{1}{\delta_{t,p}} \bigg].
\label{eqn:phiDelta}
\end{align}

$\Delta_{\chi}$ used in equation~\ref{eqn:chiUp} for $\chi$ update

\begin{align}
\Delta_{\chi^{'}_{t,p,g,h}} &= \bigg[ \Psi(\tau_{k,w_{t,p,i}}) - 
\Psi(\sum_{w_{t,p,i}} \! \tau_{k, w_{t,p,i}}) \bigg]
\nonumber \\  & + \ln \! (\frac{\epsilon}{\delta{p,t}}) \frac{1 - 
\sum_{q,h} \! \phi_{t,p,q,k,j}}{\delta_{t,p}} 
\nonumber\\  & + \frac{\sum_{q,h} \phi_{t,p,q,k,h}}{\delta_{t,p}} 
\ln \! (1 + \frac{\epsilon}{\delta_{t,p}}).
\label{eqn:chiDelta}
\end{align}

Partial derivative of $\nu$
\begin{align}
\frac{dL}{\partial\nu_{g,h}} &= \sum_{t} \! \sum_{p,q} \! 
\phi_{t,p,q,g,h} (y_{t,p,q} \Psi'(\nu_{g,h}) - \lambda_{g,h})
\nonumber\\  & + (\kappa_{g,h} - \nu_{g,h})\Psi'(\nu_{g,h}) + 
1 - \frac{\lambda_{g,h}}{\theta_{g,h}}.
\label{eqn:partialNu}
\end{align}

The traditional variational updates for the global parameters

\begin{align}
\gamma_{p,k} &= \alpha_{k} + \sum_{t} \! \sum_{q} \! \sum_{h} \! \phi_{t,p,q,k,h} 
+ \sum_{t} \! \sum_{q} \! \sum_{g} \! \phi_{t,q,p,g,k}.
\label{eqn:gammaUp}
\end{align}

\begin{align}
\nu_{g,h}^{t+1} &= \nu_{g,h}^{t}+\rho_\nu \frac{dL}{\partial\nu_{g,h}}. 
\label{eqn:nuUp}
\end{align}

\begin{align}
\lambda_{g,h} &= \frac{\bigg( \sum_{t} \! \sum_{p,q} \! \phi_{t,p,q,g,h} y_{t,p,q} + 
\kappa_{g,h} \bigg) }{
 \bigg( \bigg( \sum_{t} \! \sum_{p,q} \! \phi_{t,p,q,g,h} \bigg) + 
\frac{1}{\theta_{g,h}} \bigg) \nu_{g,h}}.
\label{eqn:lambdaUp}
\end{align}

\begin{align}
\tau_{p,v} = \nu_{v} + \sum_{t} \! \sum_{p \in t}
\bigg(\sum_{w_{t,p,i}=v}^{N_{t,p}} \chi_{t,p,i,k} \bigg).
\label{eqn:tauUp}
\end{align}

where $\rho_\nu$ is $\nu$'s gradient ascent update step-size using its partial
derivative $\frac{dL}{\partial\nu_{g,h}}$ define in equation~\ref{eqn:partialNu}.

\end{document}